\documentclass{article}

\usepackage[preprint]{neurips_2026}

\usepackage{microtype}
\usepackage{graphicx}
\usepackage{subcaption}
\usepackage{booktabs} 
\usepackage{float}
\usepackage{hyperref}

\usepackage{amsmath,amsfonts,bm}

\def\eqref#1{equation~\ref{#1}}

\def\1{\bm{1}}

\DeclareMathAlphabet{\mathsfit}{\encodingdefault}{\sfdefault}{m}{sl}
\SetMathAlphabet{\mathsfit}{bold}{\encodingdefault}{\sfdefault}{bx}{n}

\usepackage{amsmath}
\usepackage{amssymb}
\usepackage{mathtools}
\usepackage{amsthm}
\usepackage{wrapfig}

\usepackage[capitalize,noabbrev]{cleveref}

\theoremstyle{plain}
\newtheorem{theorem}{Theorem}[section]

\newtheorem{corollary}[theorem]{Corollary}
\theoremstyle{definition}

\newtheorem{assumption}[theorem]{Assumption}
\theoremstyle{remark}
\newtheorem{remark}[theorem]{Remark}

\usepackage[disable,textsize=tiny]{todonotes}

\usepackage{enumitem}

\title{Token-Level Credit Assignment via In-Context Value Learning}

\author{
  Rohin Manvi\thanks{Equal contribution. Corresponding authors.} \\
  UC Berkeley \\
  \texttt{rohinm@berkeley.edu}
  \And
  Joey Hong\footnotemark[1] \\
  UC Berkeley \\
  \texttt{joey\_hong@berkeley.edu}
  \And
  Kevin Frans \\
  UC Berkeley
  \And
  Sergey Levine \\
  UC Berkeley
}

\begin{document}

\maketitle

\begin{abstract}
Reinforcement learning has proven extraordinarily effective at fine-tuning large language models to align to human preferences and elicit reasoning capabilities.
However, when training policies for long-horizon tasks, a sparse outcome reward provides poor credit assignment, resulting in unstable and inefficient learning.
In such settings, value functions are a natural tool for providing token-level training signals and stabilizing learning.
Our work addresses how to learn value functions that are powerful enough to perform such token-level credit assignment.
Our key insight is that value functions make strong usage of \textit{privileged information}, which can be provided in-context to base language models.
In our approach, we allow the value function access to multiple auxiliary rollouts and their rewards, such that the model can leverage counterfactual futures and semantic structure to produce more meaningful value estimates. We show that value function accuracy scales with the number of in-context rollouts. At sufficient scale, this unlocks accurate per-token advantage estimates: orders of magnitude denser credit assignment than outcome reward alone.
These value estimates can then be combined with an off-the-shelf policy improvement method to perform fine-tuning.
Empirically, we demonstrate that value functions learned via our approach perform more accurate token-level credit assignment and accordingly result in faster policy improvement. Especially motivating, we find that value functions can be reliably pretrained on \textit{offline} data, and then later used for rapid adaptation from limited online samples by providing lower variance targets for policy learning.

\end{abstract}

\section{Introduction}

Reinforcement learning (RL) has been applied to large language models (LLMs) with great success, in both aligning LLMs generations to human preferences~\citep{ouyang2022training}, and eliciting strong reasoning capabilities~\citep{openai2024openaio1card,deepseekai2025deepseekr1incentivizingreasoningcapability,kimiteam2025kimik2openagentic}.
The core of RL fine-tuning involves directly optimizing a LLM policy to maximize a desired reward, such as the correctness of math or coding solutions, typically utilizing online policy gradient methods ~\citep{schulman2017proximal,shao2024deepseekmathpushinglimitsmathematical}. 

In long-horizon tasks, the efficacy of RL fine-tuning is typically bottlenecked by poor credit assignment. 
In contrast to sparse outcome rewards, a \textit{value function} (an estimator of return from any state) can provide an informative training signal over all tokens in a long rollout.
In classical RL, policy optimization over such value functions via methods such as Q-learning or actor-critic can yield significantly more sample-efficient policy learning than naive policy gradient approaches~\citep{mnih2013playing,haarnoja2018soft}.
However, in RL fine-tuning of LLMs, most algorithms utilize a learned value function, but as a baseline for observed rewards to stabilize advantage estimates~\mbox{\citep{schulman2017proximal}}, and more recent methods even drop the value function entirely, instead obtaining advantage estimates from group-based statistics ~\citep{shao2024deepseekmathpushinglimitsmathematical}.

\begin{figure*}[t]
    \centering
    \includegraphics[width=\linewidth]{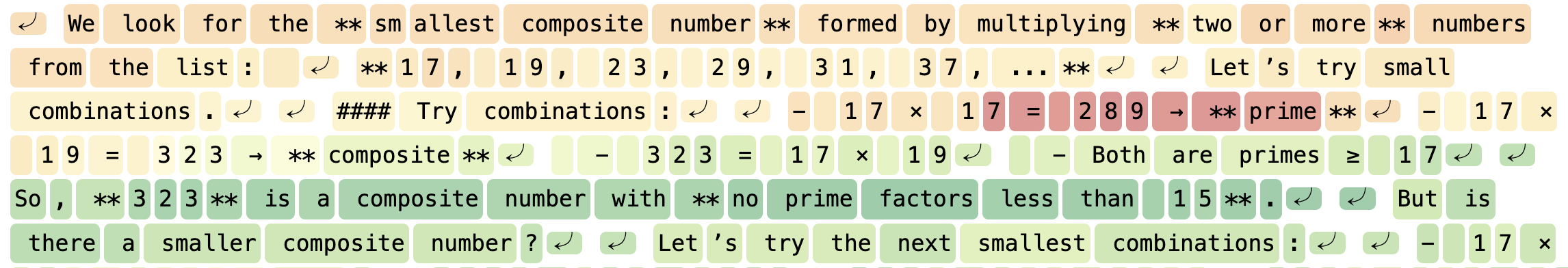}
    \caption{\textbf{Token-level value functions allow policies to improve \textit{specific} parts of their reasoning process.} Shown above, the model initially makes a mistake, but later finds the correct answer. The learned value function is able to penalize only the mistake, while incentivizing the correction steps. In contrast, commonly used outcome-only methods (e.g., GRPO) assign uniform credit to the \textit{entire trajectory}, preventing fine-grained adjustments.}
    \label{fig:teaser}
\end{figure*}

Why, then, are value functions not as useful in language modeling tasks as they are in classical RL? 
While strong value functions have the potential to drastically improve the sample-efficiency of learning policies, they must have accurate predictive quality to do so.
More precisely, accurate value prediction in these settings requires solving two distinct problems: the critic must (1) know what constitutes a high-reward trajectory, as well as (2) model the policy's future behavior.
In language modeling, the underlying token-level MDP ~\citep{ramamurthy2023is} has an enormous state (arbitrary-length token sequences) and action (token vocabulary) space, while rewards are typically sparse. 
This results in remarkably little supervision per sampled trajectory, since intermediate states receive nearly identical supervision despite differing substantially in semantic progress.
As a result, value functions trained in the standard manner are often noisy and of limited utility.

We aim to answer: \emph{what ingredient is missing from the standard value-learning recipe, and do such improvements result in stronger policy improvement?}
A key insight is that unlike policies, value functions have almost no restrictions on utilizing \textit{privileged information}, and likely greatly benefit from doing so.
As a motivating example, it is easy to see that providing a value prediction model with multiple trajectories and their associated rewards will enable it to better model the reward function, and thus make more accurate value predictions. With enough rollouts, the mean reward gives a free estimate of the initial value $V(s_0)$. 
As the value function is only used during training, we can freely provide this information without affecting the behavior of the policy at test-time. 
Thus, we argue that RL methods should take advantage of this asymmetry to the largest practical degree, especially when traditional value learning is difficult.

There is a natural strategy that arises when we note that LLMs possess strong in-context learning (ICL) capabilities, i.e., the ability to generalize from just a handful of in-context examples~\citep{garg2023transformerslearnincontext}. Rather than treating ICL as a purely test-time technique, we instead propose leveraging ICL for additional supervision during value learning. 
Specifically, when performing token-level credit assignment of rollouts by a policy, we provide the value function with \textit{multiple} counterfactual rollouts from the same problem and their observed rewards.
While policies, by construction, do not know in advance their semantic behavior, value functions used only during training can be given hindsight in the form of counterfactual futures.
By observing multiple rollouts and their outcomes, the critic can infer the general behavior of the policy and how such behavior correlates with success, enabling substantially more informative credit assignment than is available from a single trajectory.
This perspective highlights the strength of providing the value function with richer, contextual information about the current state.

Experimentally, we show that in-context value functions can be trained entirely from \textit{offline data} to perform token-level credit assignment, using auxiliary rollouts and their rewards as privileged information. By combining this value function with an off-the-shelf policy gradient method, we are able to rapidly improve the base policy using only a small amount of online data. Importantly, in-context value learning methods outperform their policy-only baselines (i.e. fine-tuning the base policy directly on offline data with RFT, DPO, or GRPO).

Our main contribution is to formulate and highlight the capabilities of in-context learning for value functions. First, we provide theoretical justification for the usage of in-context information towards more accurate value estimation. Next, we propose a practical algorithm instantiating these ideas, empirically demonstrating the benefits of in-context conditioning. Finally, we demonstrate the strength of such value functions when paired with off-the-shelf reinforcement learning methods, finding a reliable improvement over prior methodology.
Notably, we demonstrate that ICVL significantly reduces the reliance on Monte-Carlo rewards in advantage estimation, enabling true token-level credit assignment and significantly faster policy improvement.

\section{Related Work}

\textbf{Reinforcement learning for language models.} 
A rich body of work has studied the application of RL to the improvement of LLMs.
The dominant paradigm is to utilize policy gradient methods ~\citep{suttonrlbook,kakade2002approximately,schulman2017proximal} to optimize some form of trajectory-level reward, such as human feedback predictors~\citep{ouyang2022training} or verifiable outcome rewards~\citep{lambert2025tulu3pushingfrontiers}.
These algorithms either use a learned value function, or simply group-based estimates~\citep{shao2024deepseekmathpushinglimitsmathematical}, to stabilize training on Monte-Carlo rewards. 
Accordingly, many prior works have focused on improvements to the policy improvement operator~\citep{yu2025dapoopensourcellmreinforcement,yue2025vapoefficientreliablereinforcement}, but keep value learning unchanged.
Instead, recent works propose process reward models to extract dense training signal from sparse reward using human annotations~\citep{lightman2023let} or LLM-as-a-judge~\citep{zheng2023judging}, however, such strategies are often task-specific and expensive to obtain.
Our work instead considers how to improve reinforcement learning via better \textit{value learning}, and is compatible with a range of off-the-shelf policy gradient methods, including particularly efficient algorithms that require near-accurate value estimation.
 
\textbf{Value functions and actor-critic.} 
In classical RL, actor-critic algorithms such as A2C and A3C \citep{mnih2016asynchronous}, and off-policy variants like SAC ~\citep{haarnoja2018soft}, use a learned value function, or critic, as the primary signal for policy improvement. 
These methods directly extract the policy that is greedily optimal with respect to the learned value function.
This has several advantages: learning signals are dense, variance is low, and training is not as negatively impacted by long-horizon trajectories. 
In fact, when accurate, policies can be learned without any on-policy sampling~\citep{kumar2020conservative,kostrikov2021offline}.
Actor-critic algorithms have been proposed for learning LLM policies to some success in robotic manipulation~\citep{qtransformer} and language modeling~\citep{snell2023offline,hong2024qsft,li2024qprobe, zhou2024archertraininglanguagemodel}. 
However, evaluation of such approaches in natural language tasks show they are often outperformed by even supervised baselines~\citep{sodhi2023effectiveness,abdulhai2023lmrl}. 
Hence, recent approaches to RLVR rarely exploit critics in this manner.
In our work, we show that value functions learned via our approach are strong enough to provide high-fidelity token-level credit assignment, and significantly increase the sample-efficiency of policy learning.

\textbf{In-context learning.} 
Unlike tabula rasa policies, language models naturally possess in-context learning (ICL) capabilities---the ability to rapidly adapt their behavior from a small number of in-context examples~\citep{garg2023transformerslearnincontext}. The prevailing explanation for this phenomenon is that transformers perform a form of implicit gradient descent or Bayesian updating in their forward pass \citep{garg2023transformerslearnincontext,bai2023transformers}.
Our work falls in a growing body of literature that leverages ICL to improve LLMs.
Reflexion~\citep{shinn2023reflexionlanguageagentsverbal} demonstrates that LLM policies can utilize past experiences in-context to generate better actions.
In reasoning tasks, STaR~\citep{zelikman2022starbootstrappingreasoningreasoning} iteratively refines LLM generation by adding prior successful reasoning traces in-context. 
Closest to our work are methods that exploit ICL by feeding privileged information to the LLM during training~\citep{chen2025retrospective,zhao2026selfdistilled}.
Such methods typically condition a ``teacher policy" on a ground-truth output in-context, then distill the teacher onto the learned policy via distribution matching. 
Unlike these works, we do not give the \textit{policy} privileged information (which can result in unstable training due to distribution shift), and instead use this information solely for the \textit{value function}.

\section{Preliminaries}

\textbf{Reinforcement learning for language models.} We formalize text generation as a finite-horizon Markov Decision Process (MDP), following \citet{ramamurthy2023is}. 
The MDP is defined by the tuple $M = (\mathcal{S}, \mathcal{A}, R, P, \gamma, H)$ over a finite vocabulary $\mathcal{V}$, where $\mathcal{S}$ is the state space, $\mathcal{A}$ is the action space, $R$ is the reward function, $P$ is the transition function, $\gamma \in [0, 1]$ is the discount factor, and $H$ is the horizon. 
Given an input prompt $x = (x_0, \hdots, x_m)$ consisting of tokens in the vocabulary $x_i \in \mathcal{V}$, the initial state is $s_0 = x$. At timestep $t$, the LLM acts as a policy $\pi(a_t|s_t)$ that outputs the probability distribution over actions $a_t \in \mathcal{V}$. The transition function $P$ deterministically appends $a_t$ to state $s_t$, yielding next state $s_{t+1} = (x_0, \hdots, x_m, a_0, \hdots, a_t)$. The episode terminates when the model emits an end-of-sequence token \texttt{<EOS>} or the length of the generated sequence reaches the horizon $H$. 
Upon termination at timestep $T$, the environment returns a terminal reward $R(x, y)$ with solution $y = (a_0, \hdots, a_T)$.
For simplicity, the discount factor is set to $\gamma = 1$ and the value of any state $s_t$ under policy $\pi$ is the expected terminal reward from that state onward, such that $V(s_t) = \mathbb{E}_\pi \left[R(x, y)\mid s_t \right]$. The overall goal is to locate a policy $\pi_\theta$ with parameters $\theta$ that maximize expected reward:
$
J(\pi_\theta) = \mathbb{E}_{x \sim \mathcal{D}, y \sim \pi_\theta(\cdot|x)} [R(x, y)]
$.

\begin{figure*}[t]
    \centering
    \includegraphics[width=\linewidth]{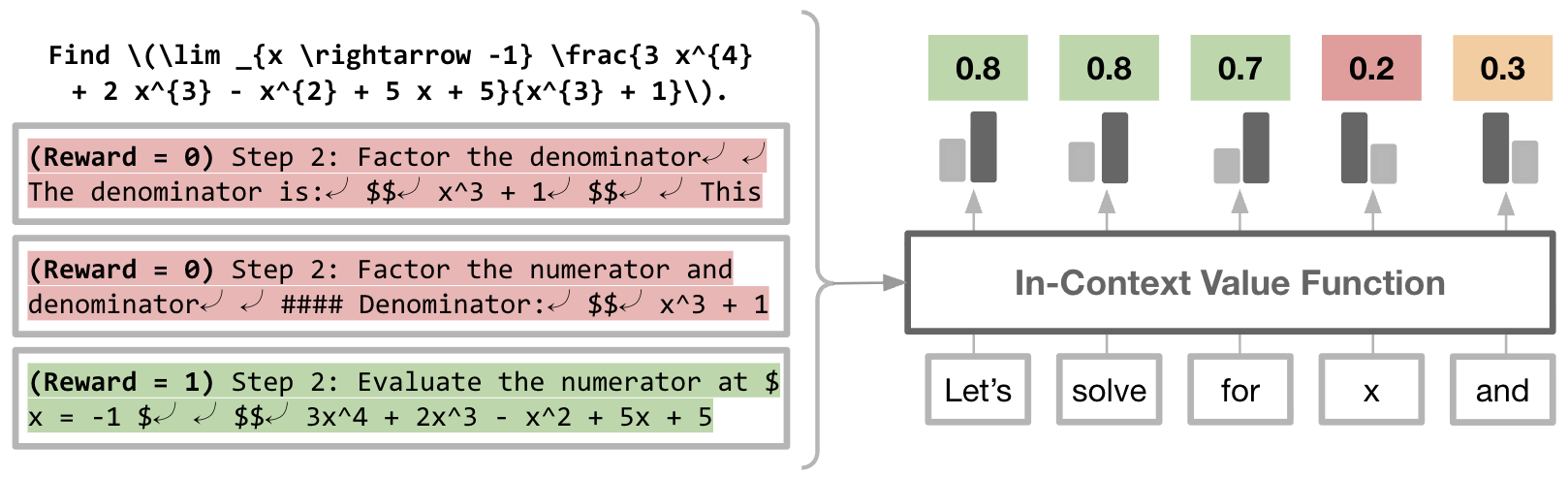}
    \caption{\textbf{The accuracy of language model value functions can be greatly improved by allowing them to condition on in-context \textit{privileged} information.} Specifically, we train value functions which condition on alternate rollouts from the same prompt, annotated with reward information. This setup provides information on 1) the reward function (via trajectories with their rewards) and 2) the reasoning behavior of the current policy, resulting in a value function that is much more robust than an unconditioned baseline.}
    \label{fig:method}
\end{figure*}

Our paper also assumes prior knowledge of policy gradient algorithms and where value functions fit in. For discussion on this, refer to Appendix~\ref{app:more_prelim}.

\section{In-Context Value Learning}

In this section, we present In-Context Value Learning (ICVL), a framework for learning high-fidelity token-level value functions by leveraging the ICL capabilities of LLMs. In later sections, we show how to practically instantiate ICVL to learn token-level value functions, and fine-tune policies in a sample-efficient manner.

We consider learning value functions from an offline dataset $\mathcal{D} = \{(x_i, y_i^{(1)}, \hdots, y_i^{(g)})\}_{i=1}^N$ of prompt-output tuples with group size of $g$ outputs per prompt,
where the outputs are generations by some policy $\pi$. 
For ease of notation, we remove index $i$ and refer to $(x, y^{(1)}, \hdots, y^{(g)})$ as a representative training sample.
Recall that the goal is to learn a value function $V(s_t) = \mathbb{E}_{\pi}[R(x, y) | s_t]$.
Note that in tasks such as mathematical reasoning, where the reward is answer correctness and lies in $\{0, 1\}$, the optimal value function is the probability of success: $V(s_t) = \mathbb{P}(R(x, y)=1 | s_t; \pi)$.

\subsection{Leveraging Privileged Information}
In standard RLVR pipelines, a value function is an LLM with parameters $\phi$ that is trained to minimize the error between the predicted value and the Monte-Carlo return:
\begin{align*}
    \arg\min_\phi \mathbb{E}_{x \sim \mathcal{D}, y \sim \pi(\cdot | x)} \left[ \sum_{t=0}^{T} \left( V_\phi(s_t) - R(x, y) \right)^2 \right]\,.
\end{align*}
In tasks with long horizons, standard value learning faces a severe supervision bottleneck: learning solely from outcome rewards is difficult because the critic must infer the following from $s_t$:
\begin{enumerate}
    \item The true \emph{reward function} $R(x, y)$, e.g., the ground-truth answer for math problems.
    \item The \emph{policy behavior} conditioned on $s_t$.
\end{enumerate}
We posit that this is prohibitively difficult to do from $s_t$ alone. So, we propose to condition the critic on privileged information $\mathcal{I}$. This privileged information should significantly reduce uncertainty over reward and policy behavior, making the value function more tractable to learn.

\emph{What would be the most effective and practical instantiation of privileged information?}
Our core insight is to leverage the in-context learning (ICL) capabilities of LLMs. 
Specifically, we let $\mathcal{I}$ consist of $g$ rollouts generated by the policy for the current prompt paired with their ground-truth rewards 
\begin{align}
\label{eq:context}
    \mathcal{I} = \{y^{(1)}, R(x, y^{(1)}), \dots, y^{(g)}, R(x, y^{(g)})\}
\end{align}
Our learning objective for the value function ends up being:
\begin{align}
\label{eq:icvl_obj}
    \arg\min_\phi \mathbb{E}_{x, \hdots, y^{(g)} \sim \mathcal{D}} \left[ \sum_{i=1}^g \sum_{t=0}^{T^{(i)}} \left( V_\phi(s^{(i)}_t, \mathcal{I}^{-i}) - R(x, y^{(i)}) \right)^2 \right]
\end{align}
where $\mathcal{I}^{-i} = \mathcal{I} \setminus \{y^{(i)}, R(x, y^{(i)})\}$ as shorthand.
Note that providing a small set of rollouts with rewards immediately anchors the value at two ends: $V(s_0)$ is the average reward across rollouts, and $V(s_T)$ at terminal states are outcome rewards. 
The critic then only needs to interpolate
for intermediate prefixes by matching them against other traces in context. This reduces the problem of inferring policy behavior to semantic pattern matching, which ICL has proven effective at.

\cref{fig:method} visualizes our proposed method. Given a prompt (left), we sample $g$ rollouts from the policy, each annotated with its binary outcome reward. When estimating the value at a token position in the target trajectory (bottom right), we condition the value function on the remaining $g-1$ rollouts and their rewards (left, colored by outcome). The model outputs per-token value estimates (top right), leveraging the in-context examples to compare the current reasoning trace against alternatives.

\subsection{Admissible Privileged Information}
\label{sec:admissible}
One might ask: why do we not include the rollout $y$ that is the completion of state $s_t$? On the surface, it would seem like this might provide the strongest hint to the value function.
However, not all choices of $\mathcal{I}$ yield a useful critic. The regression objective in \cref{eq:icvl_obj} converges to the conditional expectation $\mathbb{E}\left[ R(x, y) \mid s_t, \mathcal{I} \right]$. For policy gradient updates, the resulting advantage $R(x, y) - V_\phi(s_t, \mathcal{I})$ preserves the gradient direction only when
\begin{align*}
    \mathbb{E}\left[ \nabla \log \pi(a_t \mid s_t) \cdot V_\phi(s_t, \mathcal{I}) \right] = 0\,.
\end{align*}
Thus, we propose the necessary condition for privileged information:
\begin{remark}
    For any state $s_t$ followed by action $a_t$, the privileged information must be conditionally independent: $\mathcal{I}  \perp a_t \mid s_t$.
\end{remark}
This permits $\mathcal{I}$ to encode latent task structure or independent rollouts from the same prompt. But it forbids $\mathcal{I}$ from containing the trajectory $y$ itself, the realized return $R(x, y)$, or any noise about the future of the trajectory. ICVL satisfies the condition by construction: $\mathcal{I}$ consists of independent rollouts and their rewards from the same prompt, informing $V$ without leaking $a_t$.

The pathological case is when $\mathcal{I}$ determines $R(x, y)$. The minimizer becomes $V_\phi(s_t, \mathcal{I}) = R(x, y)$, the advantage is zero at every token, and no learning signal remains. We verify this empirically in \cref{table:admissibility}: a critic conditioned on the rollout and its reward achieves zero training loss but zero held-out per-token correlation against ICVL.

\subsection{Theoretical Analysis}
In this section, we analyze why conditioning on in-context rollouts improves value estimation. We model the value learning problem through the lens of Bayesian Linear Regression: the in-context rollouts refine the value estimate from standard learning by reducing uncertainty.

\paragraph{Setup.} First, we consider some feature representation of state $\psi(s) \in \mathbb{R}^d$. In practice, such states could be the learned representations by an LLM. In our analysis, we assume that the true value function is a linear function of these features: 

\begin{assumption}
\label{assump:linear}
The true value of a state $s$ is linear in $\psi(s)$ in terms of parameters $\phi^* \in \mathbb{R}^d$:
\begin{align*}
    V(s) = \psi(s)^\top \phi^* + \epsilon, \quad \epsilon \sim \mathcal{N}(0, \sigma^2)
\end{align*}
where $\sigma^2$ represents noise due to stochasticity of completions by the policy.
\end{assumption}
Note that this assumption is no stronger than assuming outcome rewards are linear. 
Formally, if reward $R(x, y)$ is linear, then the value $V(s)$ at any intermediate state is also linear w.r.t. $\psi(s)$, known as \textit{successor features}~\citep{barreto2018successorfeaturestransferreinforcement}.

\paragraph{Main result.}
Our main result is that additional in-context conditioning results in lower prediction error on test state $s_{test}$ for some new prompt $x_{test}$. Crucially, note that $\mathcal{I}$ contains stochastic completions from the same prompt. We defer a full proof to Appendix~\ref{app:proof}. This analysis is meant as intuition under linear-Gaussian assumptions, not as a guarantee for nonlinear Transformer critics.

\begin{theorem}
Under Assumption~\ref{assump:linear} and the idealized Bayesian in-context predictor of Appendix~\ref{app:proof}, the expected prediction error on state $s_{test}$ of a value function learned via ICVL is no higher than that learned via standard value learning, and is strictly lower whenever $\Psi_{\mathcal{I}} \Sigma_0 \psi(s_{test}) \neq 0$ (with $\Psi_{\mathcal{I}}$ the feature matrix of the context rollouts and $\Sigma_0$ the prior covariance, as in Appendix~\ref{app:proof}), i.e., whenever the context rollouts carry information correlated with the value at $s_{test}$.
\label{thm:icvl}
\end{theorem}

In summary, standard value learning must maintain a high-uncertainty estimate to cover every possible prompt and policy behavior. 
This results in a value function that is imprecise for any specific problem.
With ICVL, the additional in-context examples force the value function to focus only on the specific reasoning patterns of the current prompt. 
This effectively collapses the high uncertainty of a standard value estimate into a more confident estimate tailored exactly to the prompt at hand.

\section{Improving Token-Level Credit Assignment}

In this section, we describe a practical recipe for implementing ICVL: pretraining a value function from a prior dataset of grouped samples.
Architecturally, the value function is instantiated with a base language model; rather than adding a new prediction head, we repurpose a small number of logits from the existing output head.

Formally, we discretize the range of rewards into $B_R$ bins with boundaries $\{r_b\}_{b=1}^{B_R+1}$. We train LLM $Z_\phi$
with parameters $\phi$ to, at every state $s$, output a categorical distribution over $B_R$ tokens, where $Z_\phi(s, \mathcal{I})(b)$ denotes the predicted probability, given state $s$ and privileged information $\mathcal{I}$, that the return $R(x, y)$ falls in the range $[r_b, r_{b+1})$ (the last bin is closed). Let $b_i^*$ denote the bin containing the realized return $R(x, y^{(i)})$. Thus, our learning objective for $Z_\phi$ becomes:
\begin{align}
    \arg\min_\phi \mathbb{E}_{x, \hdots, y^{(g)} \sim \mathcal{D}} \left[ -\sum_{i=1}^g \sum_{t=0}^{T^{(i)}} \log Z_\phi(s^{(i)}_t, \mathcal{I}^{-i})(b_i^*) \right].
\end{align}
Finally, a scalar value can be extracted by marginalizing over the distribution:
\begin{align*}
V_\phi(s_t, \mathcal{I}) &= \sum_{b=1}^{B_R} \frac{r_b + r_{b+1}}{2}\; Z_\phi(s_t, \mathcal{I})(b)\,.
\end{align*}
In mathematical reasoning tasks, where the reward is binary and given by $R(x, y) \in \{0, 1\}$, we have $Z_\phi(s, \mathcal{I}) = \mathbb{P}(R(x, y) = 1 | s, \mathcal{I})$ as the output of a sigmoid function and trained via binary cross-entropy loss.

\textbf{Training Details.} We construct our training corpus by combining DeepScaleR~\citep{deepscaler2025}, the MATH training split~\citep{hendrycksmath2021}, and the GSM8K training split~\citep{cobbe2021training}. We utilize Qwen3-1.7B as the base model and generate 32 rollouts per prompt with a maximum generation length of 32k tokens (temperature 1.0, min\_p 0.1).
Each rollout is labelled with an outcome reward of either 0 or 1. For value function training, we use AdamW with a cosine learning rate schedule (5\% warmup, learning rate 1e-4, batch size 8, 1 epoch). We employ packed supervision, training the value function to predict values for 24 of the 32 in-context trajectories simultaneously.

\paragraph{Does the privileged information improve value learning?}
\begin{figure}[t]
    \centering
    \begin{subfigure}[t]{0.49\linewidth}
        \centering
        \includegraphics[width=\linewidth]{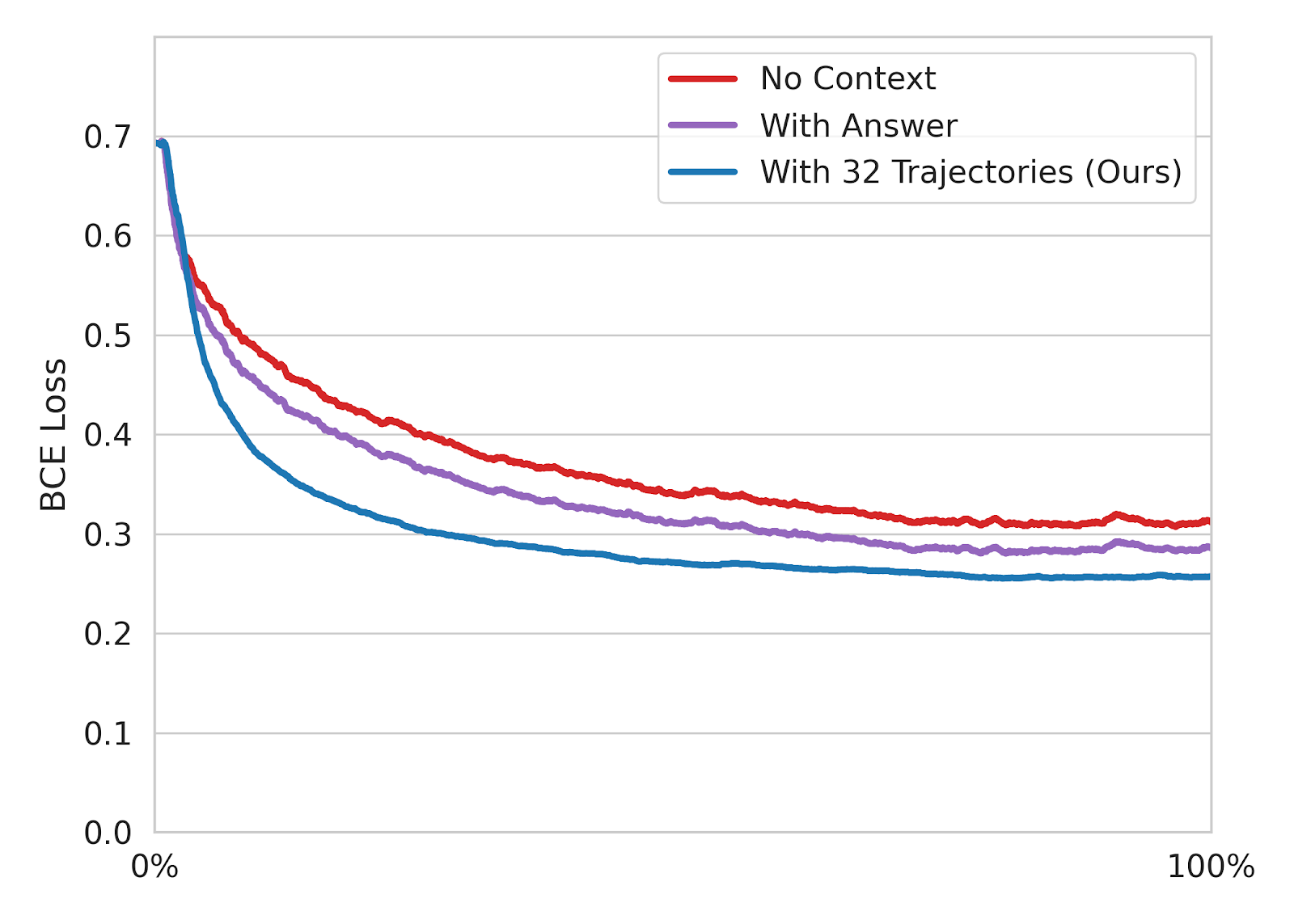}
        \caption{\textbf{Counterfactual rollouts improve value learning convergence.} We compare the (smoothed) training loss of our proposed ICVL method (conditioning on 32 trajectories) against baselines. \emph{No Context} represents naive value learning from state $s_t$ alone, while \emph{With Answer} includes the ground-truth solution in-context. We observe that accessing multiple counterfactual rollouts (\emph{With 32 Trajectories}) significantly accelerates learning and lowers the final loss compared to having access to only the true answer.}
        \label{fig:training_loss}
    \end{subfigure}
    \hfill
    \begin{subfigure}[t]{0.49\linewidth}
        \centering
        \includegraphics[width=\linewidth]{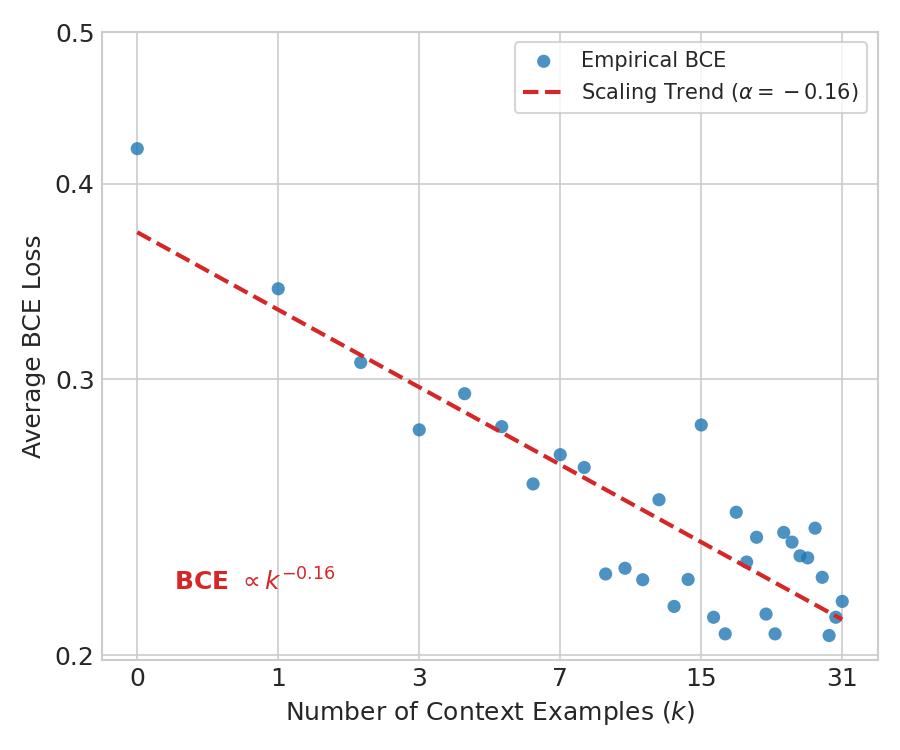}
        \caption{\textbf{Prediction quality increases as more rollouts are provided in-context.} We observe a power-law relationship between the number of in-context rollouts $k$ and binary cross-entropy (BCE) loss of the value function. The scaling exponent $\alpha \approx -0.16$ indicates consistent improvement in credit assignment accuracy as the critic is provided with more counterfactual rollouts as privileged information.}
        \label{fig:context_scaling}
    \end{subfigure}
\end{figure}

We compare the training loss of our value function when three types of contextual information are provided, each testing a different hypothesis about what information is needed for accurate value prediction:
\begin{itemize}
    \item \textbf{No Context}: Only the problem and tokens from the current rollout are provided. This is standard value learning, where the model must infer both the reward function and policy behavior from $s_t$ alone.
    \item \textbf{With Answer}: The verifiable ground-truth answer (e.g., a math expression) is provided to the value model, as in prior works~\citep{chen2025retrospective,zhao2026selfdistilled}. This tells the model how to be a perfect reward model (for math, the answer is sufficient to verify any trajectory), but provides no information on policy behavior.
    \item \textbf{With Counterfactual Trajectories} (ours): A set of reward-annotated rollouts from the same problem are provided as context. This gives the model anchor points: the value of the initial state is simply the average reward, and final states have value equal to their rewards. It also reveals policy behavior and enables semantic generalization to partial outputs.
\end{itemize}

As shown in \cref{fig:training_loss}, our proposed in-context conditioning method results in consistently lower loss across training than the \emph{No Context} or \emph{With Answer} settings. Notably, this is not just a training efficiency gain, but the asymptotic performance of the in-context value function remains stronger at convergence. This finding highlights that providing counterfactual rollouts to the value function provides information that is fundamentally unpredictable from the prompt alone.

\paragraph{How does learning scale with trajectories in the privileged information?}

We next aim to investigate how the quality of the privileged information impacts value learning, by varying the number of in-context trajectories.
We train a single value function capable of conditioning on up to 32 in-context trajectories. At evaluation, we measure validation loss on the AMC 2023 and AIME 2025 benchmarks while varying the number of trajectories provided in-context. As shown in \cref{fig:context_scaling}, we observe a power-law relationship between context size and predictive error, with scaling exponent $\alpha \approx -0.16$. This indicates that credit assignment accuracy improves consistently as more counterfactual rollouts are provided, though with diminishing returns.

\paragraph{Can the privileged information enable predicting token-level values?}

\begin{wraptable}{r}{8cm}
    \centering
    \begin{tabular}{c c c c}
        Method & Early Error & Mid Error & Late Error \\ \hline
        No Context & $0.17$ & $0.29$ & $0.05$ \\
        Ours & $\mathbf{0.08}$ & $\mathbf{0.12}$ & $\mathbf{0.02}$ \\ \hline
    \end{tabular}
    \caption{\textbf{ICVL enables token-level value learning}. We show that ICVL is able to predict values of tokens in early and middle parts of a trajectory much more accurately.}
    \label{tab:token_error}
\end{wraptable}
While we have established that privileged information can improve value learning, we have yet to demonstrate that it is sufficient for \emph{token-level} values. Namely, learning accurate values for tokens early in a trajectory is notoriously difficult due to the amount of aleatoric noise that exists. Hence, we evaluate the efficacy of our method at token-level value prediction by predicting the value of $9$ uniformly spaced states across $16$ random trajectories in the dataset. For each of $9$ states, we compute the ground-truth token-level value by averaging the mean reward achieved by branching from the state and sampling $8$ independent rollouts. We condition the squared error on whether the state falls in the first, middle, or last third of the trajectory. As shown in \cref{tab:token_error}, our approach significantly reduces the prediction error by over $50\%$ across states in a trajectory. Perhaps more importantly, we do not see a large regression in predictive quality for states in the middle of a trajectory, which are likely hardest to predict (since early and late states have too much or little noise to be interesting to predict, respectively).

Furthermore, \cref{sec:qualitative} provides four trajectories where the value transitions at specific semantic events: a self-correction, a strategy pivot, a wrong-formula commit, and an arithmetic slip. This serves as qualitative evidence that token-level values are being learned.

\paragraph{How important is maintaining independence?}

In \cref{sec:admissible} we argue that conditioning on information that determines $R(x, y)$ collapses the value function to a constant predictor. We test this by training three value functions on identical distillation targets, varying only the privileged input, and measuring training KL and held-out per-token correlation against the ICVL critic.
\cref{table:admissibility} confirms this. The trajectory-and-reward critic reaches zero training loss with zero held-out correlation, validating the prediction in \cref{sec:admissible}. The other two critics, trained on identical low-variance targets, recover only partial correlation, indicating that the remaining error comes from missing information (epistemic uncertainty) rather than noisy targets.

\begin{table}[b]
    \centering
    \begin{tabular}{lcc}
    \toprule
    Privileged information $\mathcal{I}$ & Train KL $\downarrow$ & Held-out V corr.\ $\uparrow$ \\
    \midrule
    None & 0.087 & 0.569 \\
    Ground-truth answer & 0.075 & 0.704 \\
    Same trajectory $y$ and reward $R(x, y)$ & 0.000 & 0.000 \\
    \bottomrule
    \end{tabular}
    \vspace{5pt}
    \caption{\textbf{Reward conditioning collapses the value function to a constant.} Conditioning on the trajectory and its reward minimizes the loss but produces zero per-token correlation against ICVL: the critic memorizes the return rather than estimating the value. Critics without this leakage recover only partial correlation despite identical targets, isolating epistemic rather than aleatoric uncertainty as the bottleneck.}
    \label{table:admissibility}
\end{table}

\section{Enabling Fast Policy Learning}
\label{sec:rl}

\begin{table*}
    \centering
    \begin{tabular}{lrrrr}
    \toprule
    Method & AIME 2025 & Math 500 & AMC 2023 & GSM8K \\
    \midrule
    Base Model & 9.48 & 72.35 & 45.55 & 80.89 \\
    RFT (w/ all offline data) & 13.02 & 72.45 & 47.81 & 81.41 \\
    DPO~\citep{rafailov2023direct} & 14.53 & 73.20 &  46.71 & 82.61 \\
    GRPO~\citep{shao2024deepseekmathpushinglimitsmathematical} & 18.96 & 79.60 & 58.05 & 84.95 \\
    OPSD~\citep{zhao2026selfdistilled} & 19.98 & 76.12 & 53.37 & 82.84 \\
    \midrule
    PPO ($\lambda=0.99$) & 17.29 & 78.90 & 55.08 & 83.76 \\
    PPO (w/ ICVL, $\lambda=0.99$) (ours) & 16.56 & 79.75 & 57.42 & 83.53 \\
    PPO ($\lambda=0$) & 11.46 & 71.50 & 47.81 & 80.38 \\
    \textbf{PPO (w/ ICVL, $\lambda=0$) (ours)} & \textbf{29.27} & \textbf{84.65} & \textbf{70.31} & \textbf{86.85} \\
    \bottomrule
    \end{tabular}
    \vspace{5pt}
    \caption{\textbf{ICVL-based methods outperform common baselines on rapid adaptation via reinforcement learning}. Notably, the performance gains surpass what is attainable by simply training further on more offline data (with RFT).
    Additionally, \textbf{ICVL enables the usage of a more aggressive policy improvement operator ($\lambda=0$)}, a strategy which fails when a naive value function is used.}
    \label{table:comparison}
\end{table*}

The previous section establishes that ICVL improves the quality of learned value functions. However, the end goal of value learning is to improve credit assignment during policy optimization. Hence, in this section, we show how the value functions learned via ICVL can enable much more sample-efficient policy learning.  

Our setup intentionally considers the \textit{rapid adaptation setting}, in which we have a sizable offline dataset of trajectories, and we want to utilize this dataset in a way such that our policy can rapidly improve within only a few fine-tuning steps. Using the same data and base model as described in the previous section, all methods are trained on 1024 online samples. Since RFT can leverage offline data, for RFT we report results when trained additionally on all available offline data. For policy training, we use AdamW with a cosine learning rate schedule (5\% warmup, learning rate 1e-5, batch size 8, 2 epochs). For PPO, the 1024 samples are one trajectory from each of 1024 prompts whose rollout groups contain both correct and incorrect answers. The critic (with or without ICVL) is trained once on the offline data and kept fixed: values for the 1024 trajectories are computed once, and PPO trains on these stored values. This is a one-shot adaptation setting rather than an iterative actor-critic loop.

We compare against numerous strong baselines aimed at improving the reasoning capabilities of LLMs:
\textbf{Rejection Fine-Tuning (RFT)} trains the policy only on successful on-policy completions. \textbf{Direct Preference Optimization (DPO)}~\citep{rafailov2023direct} separates positive and negative completions by outcome and increases likelihood of positives over negatives. \textbf{Group-Relative Policy Optimization (GRPO)}~\citep{shao2024deepseekmathpushinglimitsmathematical} normalizes outcome rewards by group statistics for each prompt and applies a scalar advantage per rollout. \textbf{On-Policy Self-Distillation (OPSD)}~\citep{zhao2026selfdistilled} distills a teacher policy conditioned on the ground-truth solution into the training policy via reverse KL as a token-level value. \textbf{Proximal Policy Optimization (PPO)}~\citep{schulman2017proximal}, our main vehicle for token-level value functions, computes advantages via the GAE estimator of Equation~\ref{eq:gae} under varying $\lambda$.
Of the evaluated methods, only PPO involves learning a value function. Hence, we consider both baseline PPO with standard value learning, as well as PPO with our proposed ICVL.
Note that our method, ICVL, is agnostic to how such values are used for policy learning. 
We use PPO as the policy learning algorithm simply due to its prevalence in RLVR literature.

\subsection{Results}
Results are shown in Table~\ref{table:comparison}. We report accuracy averaged over 32 independent runs for the smaller benchmarks (AIME 2025, AMC 2023) and 4 runs for larger benchmarks (Math 500, GSM8K). Among the baselines, GRPO achieves the strongest overall performance, consistent with its status as the current standard for reasoning fine-tuning.
Despite RFT being trained with all available offline data, PPO with ICVL and $\lambda=0$ reliably outperforms all baselines across each benchmark. On AIME 2025, our method achieves 29.27\% compared to GRPO's 18.96\%, a relative improvement of over 50\%. Similar gains are observed on AMC 2023 (70.31\% vs 58.05\%) and Math 500 (84.65\% vs 79.60\%). This demonstrates that accurate token-level credit assignment, enabled by ICVL, yields substantial sample efficiency gains, particularly in the fast adaptation regime. Furthermore, even though OPSD also leverages privileged information in-context, it performs noticeably worse than our approach, implying that privileged context is better used by the value function than policy itself.

\paragraph{Do stronger value functions enable more efficient policy improvement algorithms?}

Recall from the GAE formula in Equation~\ref{eq:gae} that parameter $\lambda$ controls how much we rely on the learned value function versus Monte Carlo returns. 
When $\lambda \approx 1$, the advantage estimate telescopes to the outcome reward, using the value function only as a variance-reduction baseline. 
Meanwhile, when $\lambda = 0$, advantage estimates are purely temporal differences $A_t = V(s_{t+1}) - V(s_t)$, and rely entirely on the value function for token-level credit assignment.

Table~\ref{table:comparison} reveals a striking pattern. 
For standard PPO, decreasing $\lambda$ from 0.99 to 0 causes performance to collapse (e.g., AIME drops from 17.29\% to 11.46\%), as the inaccurate value function introduces high bias. However, for PPO with ICVL, the opposite occurs: performance \emph{improves} dramatically when decreasing to $\lambda = 0$ (AIME rises from 16.56\% to 29.27\%). 
This demonstrates that ICVL produces value functions accurate enough to trust completely for credit assignment, without relying on Monte-Carlo rewards.
Also note that at $\lambda = 0.99$, ICVL provides little benefit over standard value learning (16.56\% vs 17.29\% on AIME). This is expected: high $\lambda$ discards the granular token-level signal in favor of Monte Carlo returns, negating the advantage of more accurate token-level values.

Note that at $\lambda = 0$, PPO closely resembles actor-critic methods in classical RL, where the critic directly drives policy improvement on an action-to-action level~\citep{peng2019advantage}. These methods are known to be highly sample-efficient, but only when the value function is sufficiently accurate. 
Our results suggest that ICVL potentially unlocks sample-efficient value-based RL approaches that were previously inaccessible due to the difficulty of token-level credit assignment. We leave the discovery of such algorithms to future work.

\section{Discussion}

In this work, we introduce In-Context Value Learning (ICVL), a framework that leverages the sample efficiency of in-context learning to enable token-level credit assignment when training language models via reinforcement learning. By conditioning the value function on a set of counterfactual rollouts and their outcomes \emph{only during training}, we are able to drastically reduce major sources of uncertainty during value learning: the ground-truth reward function, and policy behavior.
We demonstrate that ICVL yields significantly more accurate token-level values than standard baselines, enabling sample-efficient policy optimization even in settings where online data is scarce. Notably, we show that using our learned value function within PPO via a GAE estimate with $\lambda = 0$ (which requires accurate token-level values) outperforms a range of strong reasoning baselines.

\textbf{Limitations and Future work.}
An inherent limitation of our method is that ICVL requires a long context window for the value function, which can increase memory requirements.
Furthermore, while we demonstrate that token-level credit assignment can be achieved, we do not claim to develop specific \textit{end-to-end RL algorithms}, and instead use an off-the-shelf policy improvement step.
We believe RL approaches exist that better leverage values learned via ICVL, potentially inspired by policy extraction approaches used in off-policy and offline RL.
Finally, our experiments use a single model (Qwen3-1.7B) on mathematical reasoning with binary rewards, in a one-shot adaptation setting with a fixed critic; validating ICVL on larger models, other reward types, and iterative online RL remains future work.

\bibliography{paper}
\bibliographystyle{plainnat}

\newpage
\appendix

\section{Preliminary Discussion of Policy Gradient and Value Functions}
\label{app:more_prelim}

\textbf{Policy gradient methods.} Standard approaches optimize this objective using policy gradients, which update the policy parameters $\theta$ in the direction of the estimated gradient,
\begin{align*}
\nabla_\theta J(\pi_\theta) = \mathbb{E}\left[\sum_t A_t \nabla_\theta \log \pi_\theta(a_t|s_t)\right]\,,   
\end{align*}
where $A_t$ is the advantage function estimating how much better action $a_t$ is compared to the average action at state $s_t$. Frequently a clipping constraint is used to prevent destructively large updates~\citep{schulman2017proximal}. 
Notably, the decision of how to estimate $A_t$ is a design choice, and numerous strategies have been used. The simplest strategy, often referred to as REINFORCE~\citep{sutton2000policy}, is to simply estimate advantage using Monte-Carlo returns.

However, it is often desirable to normalize these returns to form lower-variance advantage estimates. Common strategies are to use statistics over the training batch \citep{huang202237}, or statistics within a group of rollouts from the same prompt \citep{shao2024deepseekmathpushinglimitsmathematical}. 
While such techniques are simple to implement, they still assign the same advantage $A_i$ to every token in trajectory $y_i$, effectively treating every step as equally responsible for the outcome.

\textbf{Learned token-level value functions.} A higher-resolution strategy is to learn an explicit neural network value function, which can be used to estimate $V(s_t)$ at every token. The typical setup is to train this value function towards predicting the Monte Carlo returns of each trajectory.
Once trained, this value function can either be used as a baseline value, or the \textit{temporal difference} of $r_t + V(s_{t+1}) - V(s_t)$ can be used as an advantage estimate.

\textbf{Generalized advantage estimation.} There is often a bias-variance tradeoff when utilizing Monte Carlo value estimates versus learned value estimates. A standard approach that balances these estimators is Generalized Advantage Estimation (GAE)~\citep{gae}, which can be understood as an exponentially weighted average of temporal differences:
\begin{align}
A_t &= \sum_{\ell=t}^{T} \lambda^{\ell - t} \left(r_{\ell} + V_\phi(s_{\ell+1}) - V_\phi(s_{\ell})\right)\,,
\label{eq:gae}
\end{align}
where in language modeling, $r_t = R(x, y)$ for $t = T$ and $0$ otherwise, and $V_\phi(s_{T+1}) = 0$. 
Notably, $\lambda$ controls the "smoothing" that is performed -- when $\lambda \approx 1$, we rely more on Monte Carlo estimation, and in the case where $\lambda \approx 0$, we fully rely on the one-step temporal differences by the value function.
In theory, $\lambda$ modulates a bias-variance tradeoff, where higher $\lambda$ relies on unbiased observed rewards, but at the cost of high variance and unstable training, and low $\lambda$ reduces variance, but at the cost of introducing bias due to inaccurate values.
As we will show in \cref{sec:rl}, training with $\lambda=0$ is dramatically sample-efficient but requires a sufficiently accurate value function to work.

\section{Qualitative Examples of Token-Level Credit Assignment}
\label{sec:qualitative}

\begin{figure}[H]
    \centering
    \includegraphics[width=0.8\linewidth]{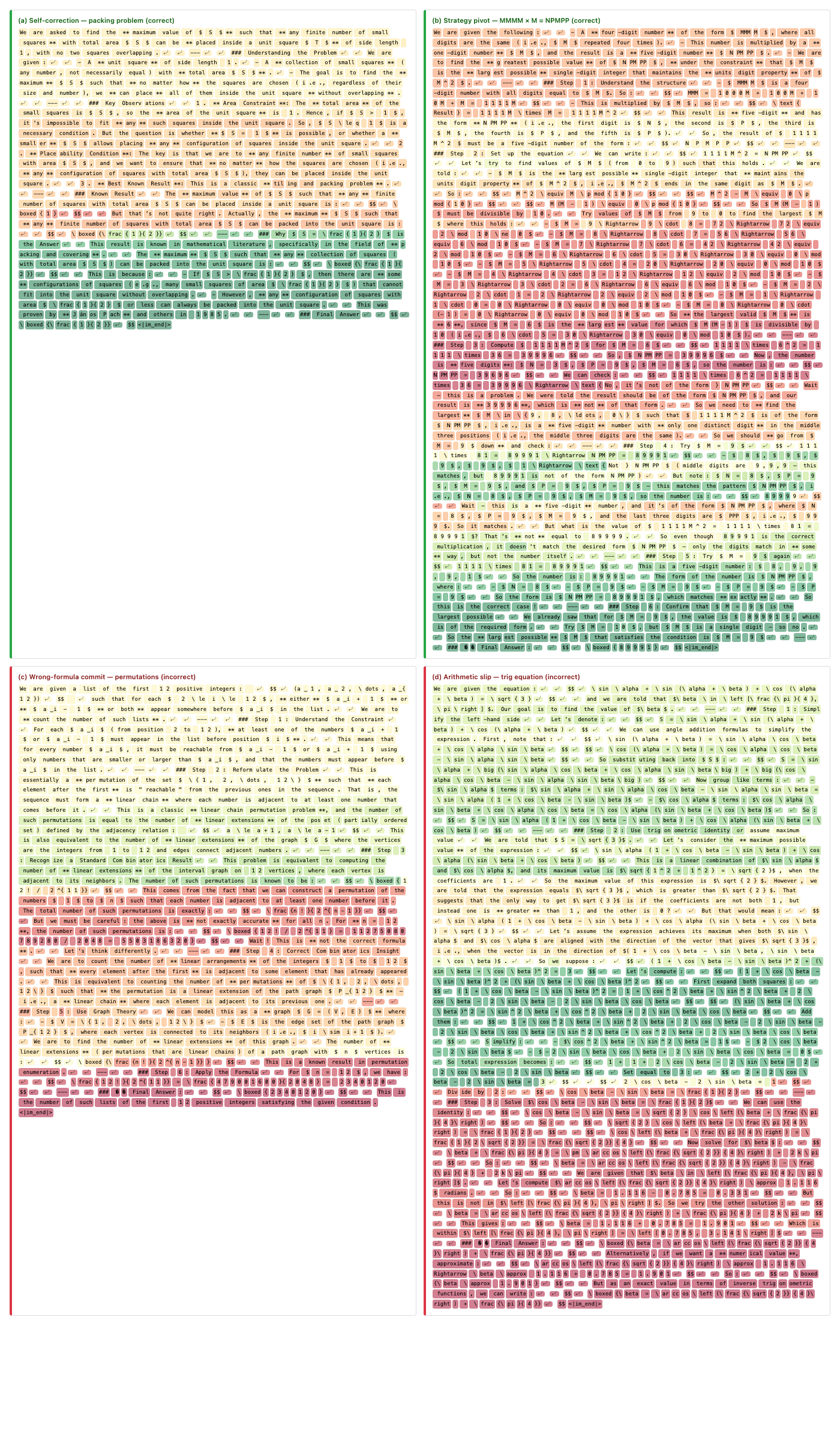}
    \caption{\textbf{Token-level value transitions trace specific reasoning events.} (a) Self-correction on a square-packing problem: $V$ rises after the model retracts $\boxed{1}$ and writes $\boxed{1/2}$. (b) Strategy pivot: model switches from $M=6$ to $M=9$ on an MMMM~$\times$~M problem; $V$ rises at the pivot. (c) Wrong-formula commit on a permutation count problem: $V$ drops at $12!/2^{11}$ before the final answer is generated. (d) Arithmetic slip on a trigonometric equation: $V$ drops at $\cos\beta - \sin\beta = 1/2$, where the correct equation equals $0$.}
    \label{fig:qualitative}
\end{figure}

\cref{fig:qualitative} shows four trajectories with per-token value estimates from the ICVL critic. Token background color reflects $V(s_t)$. Two trajectories are correct: one shows a self-correction (model writes a wrong answer, retracts, writes the correct one), one shows a mid-rollout strategy pivot. Two are incorrect: in both, $V$ drops at the wrong-decision token before the final answer is generated.

\newpage

\section{Proof of \texorpdfstring{\cref{thm:icvl}}{Theorem 4.3}}
\label{app:proof}

Let standard offline value learning on dataset $\mathcal{D}$ yield the weights $\phi_{0}$ (which we can obtain via least-squares regression).
Note that the estimate carries epistemic uncertainty due to finite data.
We parameterize this uncertainty as covariance matrix $\Sigma_0$.

Recall that in ICVL, the critic receives privileged information $\mathcal{I} = \{(y^{(i)}, R(x, y^{(i)}))\}_{i=1}^g$ consisting of $g$ rollouts from policy $\pi$, and must regress $V(s)$ for some new prefix $s$.
Motivated by \citet{garg2023transformerslearnincontext}, who find that Transformers trained on linear regression tasks closely match the Bayes-optimal predictor, we analyze an idealized critic that performs the exact Bayesian update:

\begin{assumption}[Idealized ICL as Bayes Predictor]
\label{lemma:bayes_update}
Given a prior mean $\phi_{0}$ and covariance $\Sigma_0 \succ 0$, and in-context samples $\mathcal{I}$, the critic updates its parameter estimates to the posterior mean $\phi_{post}$ and covariance $\Sigma_{post}$, following the standard Bayesian update rule:
\begin{align*}
    \phi_{post} = \Sigma_{post} \left( \Sigma_{0}^{-1} \phi_{0} + \frac{1}{\sigma^2} \Psi_{\mathcal{I}}^\top \mathbf{r}_{\mathcal{I}} \right)\,, \quad
    \Sigma_{post} = \left( \Sigma_{0}^{-1} + \frac{1}{\sigma^2} \Psi_{\mathcal{I}}^\top \Psi_{\mathcal{I}} \right)^{-1}
\end{align*}
where $\Psi_{\mathcal{I}} \in \mathbb{R}^{g \times d}$ is the matrix of feature vectors from the context rollouts, and $\mathbf{r}_{\mathcal{I}} \in \mathbb{R}^{g}$ is a vector of observed rewards. 
\end{assumption}

\begin{corollary}
The posterior covariance after ICL satisfies $\Sigma_{post} \preceq \Sigma_{0}$
\label{cor:contraction}
\end{corollary}
\begin{proof}
The update in $\Sigma_{post}$ adds the positive semi-definite matrix $\frac{1}{\sigma^2} \Psi_{\mathcal{I}}^\top \Psi_{\mathcal{I}}$ to the precision matrix. 
\end{proof}

Now, we have the expected prediction error of value estimate $\hat{V}(s_{test})$ decomposes into two sources of error:
\begin{align*}
    \mathbb{E} [ (V(s_{test}) - \hat{V}(s_{test}))^2 ]
    &= \sigma^2 + \underbrace{\psi(s_{test})^\top \text{Cov}(\phi) \psi(s_{test})}_{\text{Epistemic Error}}
\end{align*}
For the standard estimator, note that we have $\text{Cov}(\phi) = \Sigma_{0}$. For ICVL, $\text{Cov}(\phi) = \Sigma_{post}$. From Corollary~\ref{cor:contraction}, we know $\Sigma_{post} \preceq \Sigma_{0}$. Therefore:
\begin{align*}
    \psi(s_{test})^\top \Sigma_{post} \psi(s_{test}) \le \psi(s_{test})^\top \Sigma_{0} \psi(s_{test})\,.
\end{align*}
For strictness, the Woodbury identity gives $\Sigma_0 - \Sigma_{post} = \Sigma_0 \Psi_{\mathcal{I}}^\top (\sigma^2 I_g + \Psi_{\mathcal{I}} \Sigma_0 \Psi_{\mathcal{I}}^\top)^{-1} \Psi_{\mathcal{I}} \Sigma_0$, so the inequality is strict if and only if $\Psi_{\mathcal{I}} \Sigma_0 \psi(s_{test}) \neq 0$,
as desired.
Importantly, the magnitude of this reduction depends on the alignment between $\psi(s_{test})$ and the row space of $\Psi_{\mathcal{I}}$. 
Since $s_{test}$ and $\mathcal{I}$ are derived from the same prompt $x_{test}$, their feature representations are highly correlated. 
Thus, the context $\mathcal{I}$ reduces variance precisely in the directions required to predict $s_{test}$ accurately.

\clearpage

\end{document}